\documentclass[10pt, a4paper]{article}
\usepackage{lrec2016}
\usepackage{graphicx}
\usepackage{color,soul}
\usepackage{multibib}
\usepackage{lingmacros}
\usepackage[utf8x]{inputenc}
\newcites{languageresource}{Language Resources}

\title{Discriminating Similar Languages: Evaluations and Explorations}

\name{Cyril Goutte\textsuperscript{1}, Serge L{\'e}ger\textsuperscript{1}, Shervin Malmasi\textsuperscript{2}, Marcos Zampieri\textsuperscript{3,4}}

\address{National Research Council (NRC), Canada\textsuperscript{1}, Macquarie University, Australia\textsuperscript{2} \\
               Saarland University, Germany\textsuperscript{3}, German Research Center for Artificial Intelligence (DFKI), Germany\textsuperscript{4}\\
          firstname.lastname@nrc.ca, shervin.malmasi@mq.edu.au, marcos.zampieri@uni-saarland.de\\}

\abstract{We present an analysis of the performance of machine
  learning classifiers on discriminating between similar languages and
  language varieties. We carried out a number of experiments using
  the results of the two editions of the Discriminating between
  Similar Languages (DSL) shared task. We investigate the progress
  made between the two tasks, estimate an upper bound on possible
  performance using ensemble and oracle combination, and provide
  learning curves to help us understand which languages are more
  challenging. A number of difficult sentences are identified and
  investigated further with human annotation.\\
\newline
\Keywords{language identification, language varieties, evaluation}}

\begin{document}

\maketitleabstract

\section{Introduction}

Discriminating between similar languages and language varieties is one of the main challenges of state-of-the-art language identification systems \cite{tiedemann12}. Closely-related languages such as Indonesian and Malay or Croatian and Serbian are very similar both at their spoken and at their written forms making it difficult for systems to discriminate between them. Varieties of the same language, e.g. Spanish from South America or Spain, are even more difficult to detect than similar languages. Nevertheless, in both cases, recent work has shown that it is possible to train algorithms to discriminate between similar languages and language varieties with high accuracy \cite{goutteetal2014,malmasi2015dsl}.

This study looks in more detail into the features that help algorithms discriminating between similar languages, taking into account the results of two recent editions of the Discriminating between Similar Languages (DSL) shared task \cite{zampieri2014vardial,zampieri15dsl}. The analysis and results of this paper complement the information presented in the shared task reports and provide novel and important information for researchers and developers interested in language identification and particularly in the problem of discriminating between similar languages.

\section{Related Work}

Language identification in written texts is a well-established research topic in computational linguistics. Interest in the task is evidenced by early n-gram-based approaches \cite{dunning94,grefenstette95} to more recent studies \cite{brown13,lui2014automatic,brown14,simoes2014}. The interest in the discrimination of similar languages, language varieties, and dialects is more recent but it has been growing in the past few years. Examples of studies include the cases of Malay and Indonesian \cite{ranaivo06}, Chinese varieties \cite{huang08}, South Slavic languages \cite{ljubesic07,ljubesic15}, Portuguese varieties \cite{zampieriandgebre12}, Spanish varieties \cite{zampierietal13,maier2014}, English varieties \cite{lui13}, Persian and Dari \cite{malmasi-dras2015}, Romanian dialects \cite{ciobanu2016}, and a number of studies on Arabic dialects \cite{elfardy13,zaidan14,tillmann-mansour-alonaizan:2014:VarDial,malmasietal2015}

A number of shared tasks on language identification have been organized in the recent years ranging from general-purpose language identification \cite{baldwin10} to more specific challenges such as the TweetLID shared task which focused on Twitter data \cite{zubiaga14,zubiaga2015tweetlid}, the shared task on Language Identification in Code-Switched Data \cite{solorio14}, and the two editions of the discriminating between similar languages (DSL) shared task. To our knowledge, however, no comprehensive analysis of the kind we are proposing in this paper has been carried out on the results obtained in a language identification shared task and our work fills this gap. The most similar analysis was applied to Native Language Identification (NLI)\footnote{This task focuses on identifying the mother tongue of a learner writer based on stylistic cues; all the texts are in the same language \cite{malmasi-lth,malmasi-mnli}.} using the 2013 NLI shared task dataset \cite{malmasi2015oracle}.

In the next sections we present the systems that participated in the two editions of the DSL shared task.

\subsection{DSL Shared Task 2014}

The first edition of the DSL task was organized in 2014 within the scope of the workshop on Applying NLP Tools to Similar Languages, Varieties and Dialects (VarDial) co-located with COLING. The organizers compiled and released a new dataset for this purpose, which they claim to be the first resource of its kind \cite{tan14}. The dataset is entitled \emph{DSL Corpus Collection}, or DSLCC, and it includes short excerpts from journalistic texts from previously released corpora and repository.\footnote{See \newcite{tan14} for a complete list of sources.} Texts in the DSLCC v. 1.0 were written in thirteen languages or language varieties and divided into the following six groups: Group A (Bosnian, Croatian, Serbian), Group B (Indonesian, Malay), Group C (Czech, Slovak), Group D (Brazilian Portuguese, European Portuguese), Group E (Peninsular Spanish, Argentine Spanish), and Group F\footnote{There were many cases of republication (e.g. British texts republished by an American newspaper and tagged as American by the original sources) that made the task for this language group unfeasible.\cite{zampieri2014vardial}} (American English, British English).

In the 2014 edition, eight teams participated and submitted results to the DSL shared task (eight teams in the closed and two teams in the open submission). Five of these teams wrote system description papers. The complete shared task report is available in \newcite{zampieri2014vardial}. We summarize the results in Table~\ref{dsl2014} in terms of accuracy (best performing entries displayed in bold).

\begin{table}[htb]
\centering
\scalebox{0.90}{
    \begin{tabular}{l c c r}
    \hline
    \bf Team       &  \bf Closed & \bf Open & \bf System Description \\ \hline
    NRC-CNRC             & \bf 95.7   & -   &  \cite{goutteetal2014}   \\
    RAE                  & 94.7   & -      & \cite{portaandsancho2014} \\
    UMich                & 93.2    & 85.9  & \cite{kingetal2014}       \\
    UniMelb-NLP         & 91.8    & \bf 88.0 & \cite{luietal2014}     \\
    QMUL                 & 90.6   & -   &  \cite{purver2014}   \\
    LIRA                 & 76.6   & -   &  -   \\
    UDE                  & 68.1   & -   &  - \\
    CLCG                 & 45.3   & -   &  -  \\ \hline
    \end{tabular}
}
\caption{DSL Shared Task 2014 - Accuracy results. Teams ranked by their results in the closed submission.}
\label{dsl2014}
\end{table}

In the closed submission track the best performance was obtained by the NRC-CNRC \cite{goutteetal2014} team, which used a two-step classification approach to predict first the language group of the text, and subsequently the language. Both NRC-CNRC \cite{goutteetal2014} and QMUL \cite{purver2014}, ranked 5\textsuperscript{th} used linear support vector machines (SVM) classifiers with words and characters as features. 

Two teams used information gain to estimate the best features for classification, UMich \cite{kingetal2014} and UniMelb-NLP \cite{luietal2014}. These two teams were also the only ones teams which compiled and used additional training material to compete in the open submission track. As can be seen in Table 1, the performance of open submissions were worse than the closed submissions. Accuracy dropped from 93.2\% to 85.9\% for UMich, and from 91.8\% to 88.0\% for UniMelb-NLP. This is probably because along with diatopic variation, systems learn properties of datasets that are often topic or genre specific. Therefore, part of what was learned from the additional training corpora was not helpful for predictions on the test set.

The RAE team \cite{portaandsancho2014} proposed an approached based on `white lists' of words used exclusively in a given language or language variety and their closed submission ranked 2\textsuperscript{nd}.

\subsection{DSL Shared Task 2015}

The 2015 edition of the DSL shared task was organized within the scope of the Joint Workshop on Language Technology for Closely Related Languages, Varieties and Dialects (LT4VarDial) co-located with RANLP.

For the DSL 2015, organizers released version 2.0 of the DSLCC which contained the same set of languages and language varieties as version 1.0 in groups A to E. The two main modifications between the two versions are the exclusion of group F (British and American English) and the inclusion of group G (Bulgarian and Macedonian).\footnote{In the 2015 edition, organizers did not use language group names as in the 2014 edition. We use them for both editions in this paper for the sake of clarity and consistency.} A new addition in the DSL 2015 is the use of two test sets (A and B). In test set A instances are presented exactly as they appear in newspaper texts whereas in test set B named entities were substituted by placeholders. According to the organizers, the release of test set B aimed to evaluate the extent to which named entities influence classification performance.

Ten teams submitted their results and eight of them published system description papers. Results of the DSL 2015 are described in detail in \newcite{zampieri15dsl} In Table 2 we summarize the results obtained by all teams using test sets A and B in both open and close submissions. The best results for each submission type are displayed in bold.

\begin{table}[!ht]
\centering
\scalebox{0.90}{
    \begin{tabular}{lccr}
   	\hline
    \bf Team   & \bf Closed & \bf Open & \bf System Description \\ \hline
    \bf & & & \bf Test Set A \\
    \hline
    MAC     & \bf 95.54 & - & \cite{malmasi2015dsl}  \\
    MMS     & 95.24 & - &  \cite{zampieri2015dsl} \\
    NRC     & 95.24 &  \bf 95.65 & \cite{goutte2015dsl}  \\
    SUKI    & 94.67 & - & \cite{jauhiainen2015dsl} \\
    BOBICEV & 94.14 & - & \cite{bobicev2015dsl} \\
    BRUNIBP & 93.66 & - & \cite{acs2015dsl}  \\
    PRHLT   & 92.74 & - & \cite{salvador2015dsl}  \\
    INRIA   & 83.91 & - & -  \\
    NLEL    & 64.04 & 91.84 & \cite{boluda2015dsl}  \\
    OSEVAL  & - & 76.17 & - \\
    \hline
    & & & \bf Test Set B \\
    \hline
       MAC     & \bf 94.01  & - & \cite{malmasi2015dsl} \\
	SUKI    & 93.02  & - & \cite{jauhiainen2015dsl} \\
    NRC     & 93.01  & \bf 93.41 & \cite{goutte2015dsl} \\
    MMS     & 92.78  & - & \cite{zampieri2015dsl} \\
    BOBICEV & 92.22  & - &  \cite{bobicev2015dsl} \\
    PRHLT   & 90.80 & - & \cite{salvador2015dsl} \\
    NLEL    & 62.78  & 89.56 & \cite{boluda2015dsl}  \\
    OSEVAL  & - & 75.30  & - \\
    \hline
    \end{tabular}
}
\caption{DSL Shared Task 2015 - Accuracy results for open and closed submissions using test sets A and B. Teams ranked by their results in the closed submission.}
\label{tab:2015normal}
\end{table}

We observe that for all systems, performance dropped from test set A (complete texts) to test set B (name entities removed). This was the expected outcome. However, performance is, in most cases, only 1 or 2 percentage points lower which is not as great as one could expect. This means that even if all place and person names, which are to a large extent country specific, are removed from Brazilian texts, systems are still able to distinguish them from, for example, Portuguese texts with very high accuracy. Another interesting aspect to observe is that, unlike in the DSL 2014, the use of additional training material helped system performance as can be observed in the NRC and NLEL submissions. This is mainly due to corpus comparability, as in 2015 team were allowed to use the DSLCC v. 1.0 and in 2014 they did not have such a resource available, and had to acquire additional, unrelated material.

The best system in the closed submission for test set A and B was MAC \cite{malmasi2015dsl} which proposed an ensemble of SVM classifiers for this task. Two other SVM-based approaches were tied in 2\textsuperscript{nd} for test set A, one by the NRC team \cite{goutte2015dsl} and MMS \cite{zampieri2015dsl}, which experimented with three different approaches and obtained the best results combining TF-IDF and an SVM classifier previously used for native language identification \cite{gebreetal2013}. The NRC team included members of NRC-CNRC, winners of the DSL closed submission track in 2014. Both in 2014 and in 2015 they used a two-stage classification approach to predict first the language group and then the language within the predicted group. Two other teams used two-stage classification approaches: NLEL \cite{boluda2015dsl} and BRUniBP \cite{acs2015dsl}.

A number of computational techniques have been explored in the DSL 2015 including token-based backoff by SUKI team \cite{jauhiainen2015dsl}, prediction by partial matching (PPM) by BOBICEV \cite{bobicev2015dsl}, and word and sentence vectors by PRHLT \cite{salvador2015dsl}. 


\section{Methods}

In the next subsections we describe the methodology behind our 4 experiments as well as the data used.

\subsection{Data}

All experiments reported here are performed on the DSL Corpus
Collection (DSLCC) versions 1.0. and 2.0. \cite{tan14}.  Both versions
cover five groups of two to three languages or varieties each (groups A-E, Table \ref{tab:data}). The 2015 collection adds Bulgarian and Macedonian (group G) plus sentences from ``Other'' languages. In a couple of experiments (Sections \ref{sec:method-ensemble} \&
\ref{sec:results-ensemble}) we use the output of the 22 entries submitted to the 2015 Shared Task.\footnote{\texttt{\small https://github.com/Simdiva/DSL-Task}.}

\begin{table}[!ht]
\centering
\small
\begin{tabular}{clrrrr}
\hline
       &    & \multicolumn{2}{c}{2014} & \multicolumn{2}{c}{2015} \\
 Grp & Language/variety   & Train & Test & Train & Test \\
 \hline
   & Bosnian             & 20k & 1000 & 20k & 1000 \\
\textbf{A} & Croatian    & 20k & 1000 & 20k & 1000 \\
   & Serbian             & 20k & 1000 & 20k & 1000 \\
\textbf{B} & Indonesian  & 20k & 1000 & 20k & 1000 \\
   & Malaysian           & 20k & 1000 & 20k & 1000 \\
\textbf{C} & Czech       & 20k & 1000 & 20k & 1000 \\
   & Slovak              & 20k & 1000 & 20k & 1000 \\
\textbf{D}
   & Brazil Portuguese   & 20k & 1000 & 20k & 1000 \\
   & European Portuguese & 20k & 1000 & 20k & 1000 \\
\textbf{E}
   & Argentine Spanish   & 20k & 1000 & 20k & 1000 \\
   & Peninsular Spanish       & 20k & 1000 & 20k & 1000 \\
\textbf{G} & Bulgarian   &    -   &  -   & 20k & 1000 \\
   & Macedonian          &    -   &  -   & 20k & 1000 \\
\hline
\textbf{X} & Others      &    -   &  -   & 20k & 1000 \\
\hline
\end{tabular}
\caption{Number of sentences in the DSLCC v. 1.0 and 2.0.}
\label{tab:data}
\label{tab:dsltask}
\end{table}

\subsection{Progress Test}
\label{sec:method-progress}
In our first experiment, we evaluate the improvements achieved from
one shared task to the other. For that purpose, we measure the
performance, on the 2014 and 2015 test data, of three systems
representative of the top performance in both years:
\begin{itemize}
\item the top 2014 system, NRC-closed-2014 \cite{goutteetal2014};
\item the top 2015 closed task system, MAC-closed-2015 \cite{malmasi2015dsl};
\item the top 2015 open task system, NRC-open-2015 \cite{goutte2015dsl}.
\end{itemize}
The 2015 shared task had two key additions: a new group of close
languages (Bulgarian/Macedonian, group G) and data from other
languages (group X). The 2015 results were measured on all groups,
but the 2014 system was not trained to recognized either group G or
group X data. As a consequence, in addition to the full 2014 and 2015
test sets, we evaluated performance on the subset of the 2015 test set
that contains the groups in the 2014 shared task, i.e. groups A to E
(5 groups and 11 variants).

\subsection{Ensemble and Oracle}
\label{sec:method-ensemble}
An interesting research question for this task is to measure the
upper-bound on accuracy.  This can be measured by treating each shared
task submission as an independent system and combining the results
using ensemble fusion methods such as a plurality voting or oracle.
This type of analysis has previously been shown to be informative for
the similar task of Native Language Identification
\cite{malmasi2015oracle}.  Moreover, this analysis can also help
reveal interesting error patterns in the submissions.

Following the approach of \newcite{malmasi2015oracle}, we apply the
following combination methods to the data.

\paragraph{Plurality Voting:} This is the standard combination strategy that selects the label with the highest number of votes, regardless of the percentage of votes it received \cite{polikar2006ensemble}. This differs from a \textit{majority} vote combiner where a label must obtain over $50\%$ of the votes.

\paragraph{Oracle:}
An oracle is a type of multiple classifier fusion method that can be
used to combine the results of an ensemble of classifiers which are
all used to classify a dataset.  The oracle will assign the correct
class label for an instance if at least one of the constituent
classifiers in the system produces the correct label for that data
point.  This method has previously been used to analyze the limits of
majority vote classifier combination \cite{kuncheva2001decision}.
It can help quantify the \textit{potential} upper limit of an
ensemble's performance on the given data and how this performance
varies with different ensemble configurations and combinations.

\paragraph{Accuracy@$N$:}
%
To account for the possibility that a classifier may predict the
correct label by chance (with a probability determined by the random
baseline) and thus exaggerate the oracle score, an Accuracy@$N$
combiner has been proposed \cite{malmasi2015oracle}
This method is inspired by the ``Precision at $k$" metric from
Information Retrieval \cite{manning:2008:eval} which measures
precision at fixed low levels of results (\textit{e.g.} the top $10$
results).
Here, it is an extension of the Plurality vote combiner where instead
of selecting the label with the highest votes, the labels are ranked
by their vote counts and an instance is correctly classified if the
true label is in the top $N$ ranked candidates.\footnote{In case of
  ties we choose randomly from the labels with the same number of
  votes.}
Another way to view it is as a more restricted version of the Oracle
combiner that is limited to the top $N$ ranked candidates in order to
minimize the influence of a single classifier having chosen the
correct label by chance.
In this study we experiment with $N = 2$ and $3$. We also note that
setting $N = 1$ is equivalent to the Plurality voting method.

Results from the above combiners are compared to a random
baseline and to the best system in the shared task.

\subsection{Learning Curves}
\label{sec:method-lc}
Learning curves are an important tool to understand how statistical
models learn from data. They show how the models behave, in terms of
performance, with increasing amounts of data. In order to compute
learning curves for the DSL task, we picked a simple model that is
easy to train and performs close to the top systems.

From the full training set, we subsample data at various sample
sizes. In order to keep the training data balanced, we sample the same
amount $N_s$ of examples from each language variant. In our setup, we
use $N_s=20,000$ (full training set), $10,000$, 5000, 2000, 1000, 500,
200 and 100. For each subsample, we train a statistical model, and test
it on the official 2015 test set. We replicate this experiment 10
times at each sample size, except for the full training set. This helps us estimate the
expected performance at each sample size, as well as error bars on the
expectation.

\subsection{Manual Annotation}
\label{sec:method-annotation}
Finally, to make this evaluation even more comprehensive, we also conducted a human evaluation experiment on some of the misclassified instances. We asked human annotators to assign the correct language or language variety of each sentence for the most difficult language groups, namely group A (Bosnian, Croatian, and Serbian), D (Brazilian and European Portuguese), and E (Argentinian and Peninsular Spanish). In this experiment we included all instances that were misclassified by the Oracle (i.e. no submission got right). For groups D and E, we added sentences that were incorrectly classified by the plurality vote method as well amounting to twelve instances per group.

Such analyses of misclassifications can provide further insights and help better understand the difficulties of the task. \newcite{acs2015dsl} showed that for 52 misclassified Portuguese instances, only 22 have been labeled correctly by the annotators with low inter-annotator agreement. We report the results obtained by the manual evaluation step in Section \ref{sec:results-annotation}.



\section{Results}


\subsection{Progress Test}
\label{sec:results-progress}
Table \ref{tab:progress} displays the results of the progress
test. The 2014 system is evaluated on the 2014 test set and 2015
progress set containing 5 groups and 11 languages. The 2015 systems
are evaluated on the full 2014 and 2015 test sets and the 2015
progress set.

\begin{table}[!ht]
\centering
\begin{tabular}{rrrr}
\hline
        & 2014 & \multicolumn{2}{c}{2015} \\
\textbf{System}  & test & progress & test \\
\hline
\textbf{NRC-closed-2014} & 95.70 & 90.15 & - \\
\hline
\textbf{MAC-closed-2015} & 96.31 & 94.33 & 95.54 \\
\textbf{NRC-open-2015}   & 96.04 & 94.48 & 95.65 \\
\hline
\end{tabular}
\caption{Progress test for 2014 and 2015 systems. ``progress'' is
  the 2015 test set without group G and X.}
\label{tab:progress}
\end{table}

Table \ref{tab:progress} shows that, when measured on the 2014 test set, the increase in
performance of the 2015 systems, although significant, is modest
(+.3-.6\%).  It should be noted however that the 2014 system is the
only one for which the training data exactly matches the test
data. Somewhat surprisingly, {MAC-closed-2015}, the best 2015
closed track submission, which is trained only on the 2015 training set,
performs slightly better than {NRC-open-2015}, which was trained
on both 2014 and 2015 data, and would therefore be expected to perform
better on 2014 data.

Year-to-year improvements are more evident from the results obtained on the
progress test. We note that average performance on the progress test is lower
than on the full 2015 test set. This is due to the omission of groups G and X, on
which 2015 systems performed very well. Looking at the performance of the best 2014 system, it is apparent that the 2015 progress set was harder than the 2014 test set. This is most likely due to the much shorter sentences providing less evidence for ngram statistics. On the progress set, the 2015 systems outperform the
2014 system by more than 4\%. This indicates that the 2014 system suffers from the mismatch in data, and suggests a large year-to-year improvement in
performance on shorter sentences.

\subsection{Ensemble and Oracle}
\label{sec:results-ensemble}

The 22 entries in the shared task (normal test set) were combined in
various ensembles and the results are shown below in
Table~\ref{tab:oracle-task}.  We observe that a plurality vote among
all the entries yields only a very small improvement over the best
single system \cite{malmasi2015dsl}.

\begin{table}[!ht]
\centering
\scalebox{1.0}{
\begin{tabular}{rc}
\hline
 & Accuracy (\%) \\
\hline
\textbf{Random Baseline}	&	$~~7.14$ \\
\textbf{Shared Task Best}	&	$95.54$ \\
\hline
\textbf{Plurality Vote}		& $96.04$ \\
\hline
\textbf{Oracle}				& $99.83$ \\
\textbf{Accuracy@3}			& $99.83$  \\
\textbf{Accuracy@2}			& $99.47$ \\
\hline
\end{tabular}
}
\caption{Ensemble results on the DSL 2015 shared task systems. The ensemble consists of all 22 submission from all teams.}
\label{tab:oracle-task}
\end{table}

The oracle results, however, are substantially higher than the voting
ensemble and close to 100\% accuracy.  The accuracy@2 and accuracy@3
results are almost identical to the full oracle, suggesting that
almost all of the errors are the result of a confusion between the top 2-3
results.
This is due to the fact that DSL errors are almost always within a
group, i.e. between 2 or 3 variants. As shown in the learning curves in Figure\ref{fig:lc}, group prediction reaches perfect performance using relatively few examples, so the remaining confusions are always within a group of languages or variants. This differs from results observed for
Native Language Identification where there is a large difference
between the oracle and accuracy@2 results.

\subsection{Learning Curves}
\label{sec:results-lc}
For simplicity, learning curves were obtained for a simple system
trained only on character 6-grams. This is essentially NRC's first 2015 run,
which performed 0.7\% below the top system, trained using various training set sizes.

\begin{figure}[!ht]
\centering
\includegraphics[width=.47\textwidth]{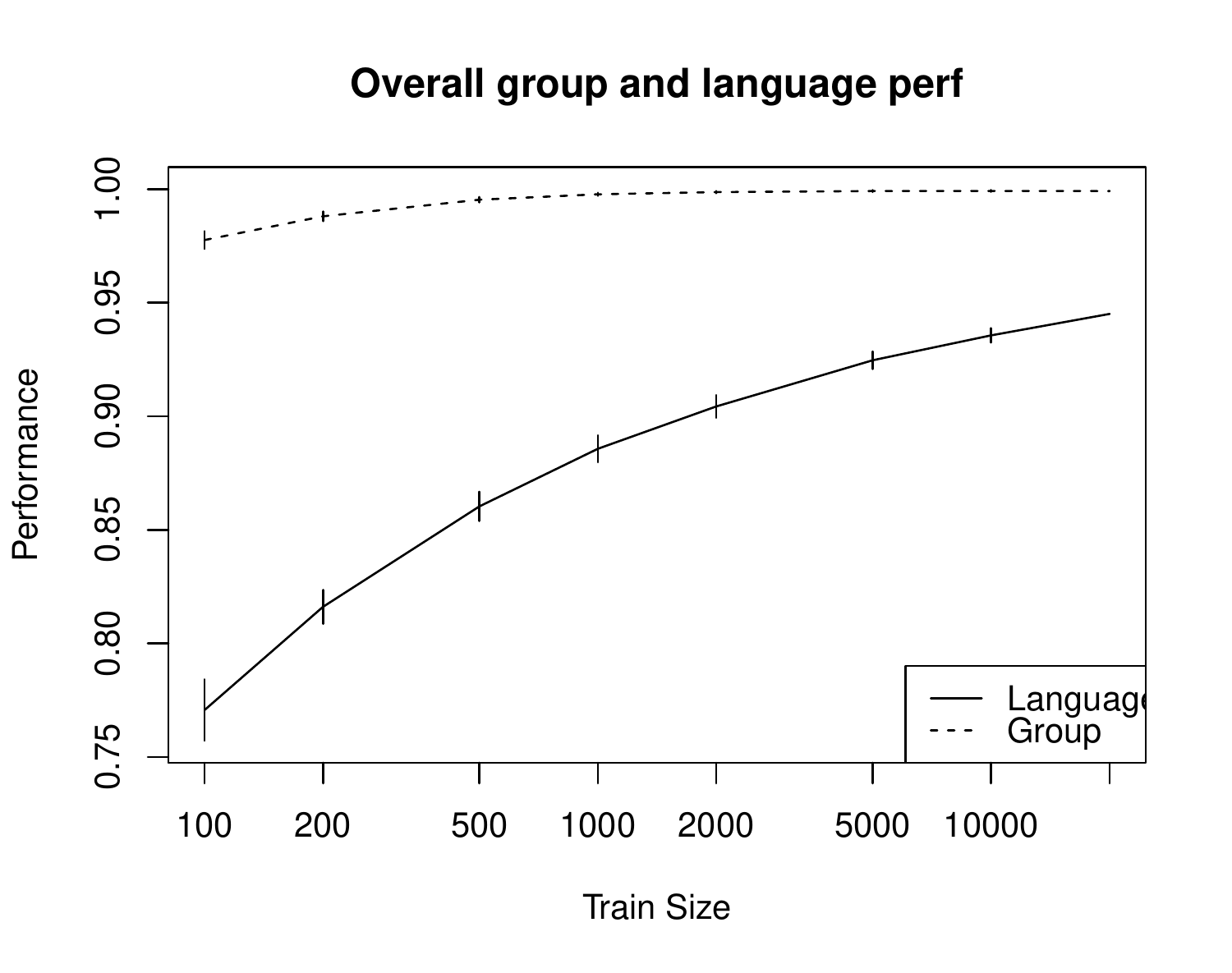}
\includegraphics[width=.47\textwidth]{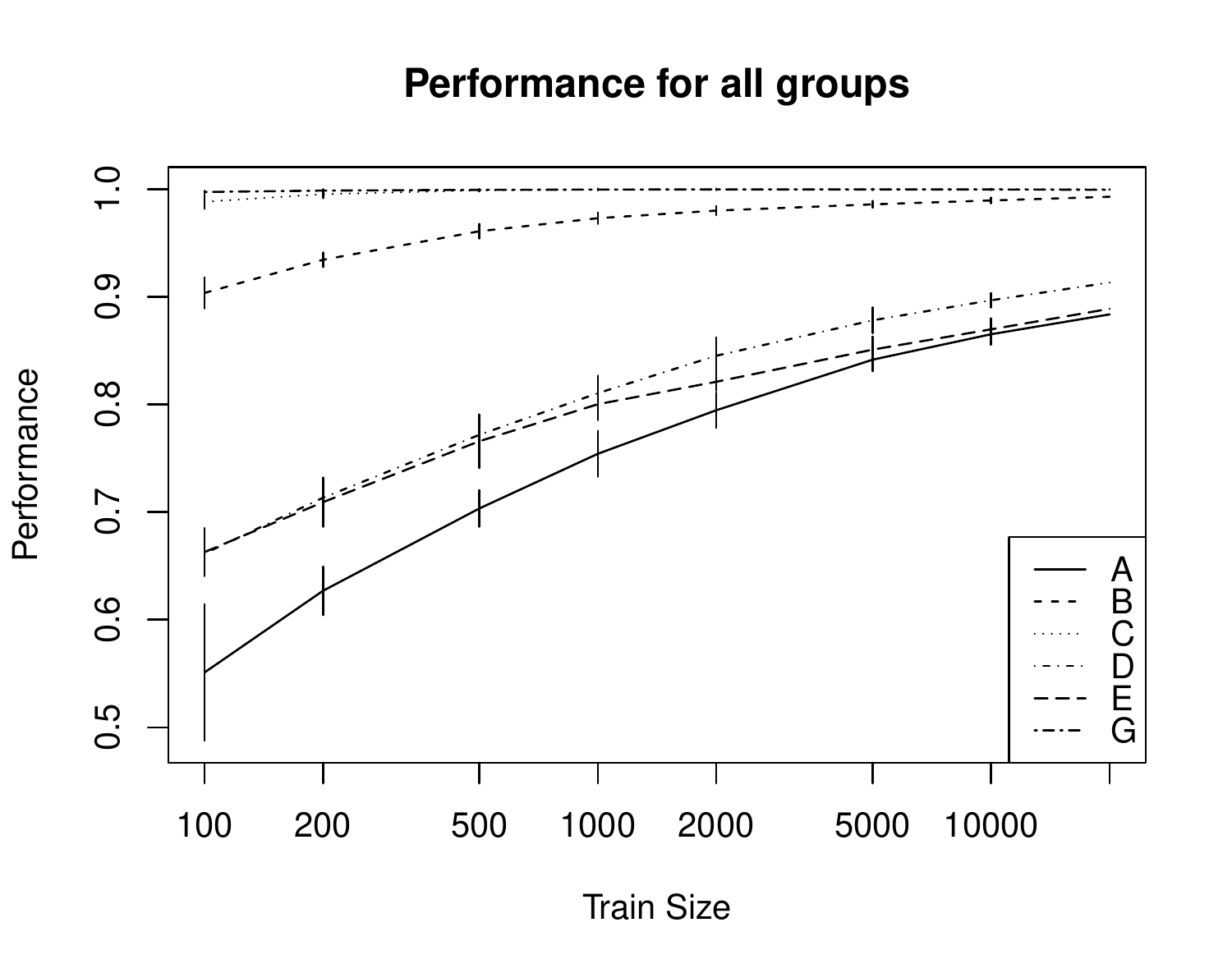}
\caption{Learning curves for average group and language
  prediction (top), and within group performance (bottom).}
\label{fig:lc}
\end{figure}

Figure \ref{fig:lc} shows learning curves for the average group and
language classification performance (top) and within each group
(bottom). The group-level curve (dashed, top) shows that predicting the group is done perfectly from around 1000 examples per language. Average language-level performance is lower and still increasing at 20k examples per language.

Looking closer at discrimination performance for each group, we see that performance for groups C, G is essentially perfect as early as 100-500 examples. This
suggests that discriminating Czech from Slovak and Bulgarian from Macedonian is easy. In those cases, it may be more challenging to investigate side issues such as
robustness to changes in source, genre, or regional proximity.  Group B was clearly
harder to learn, but there is little room for improvement above $\sim$10,000
examples/language.

Groups A, D and E display more interesting learning curves. Learning
still takes place for the full training set. The rate of progress
slows down, but doesn't seem to plateau. This is reflected on the
average performance: the last doubling of the
training data (right of the curves) brings around 1.6-1.9 additional percent of accuracy, while the first brought 4.6-7.6\% increase in performance. This suggest that, despite diminishing returns, bringing more data covering languages in these three groups would still improve prediction accuracy.

\subsection{Manual Annotation}
\label{sec:results-annotation}
As noted in Section \ref{sec:results-ensemble}, the oracle achieves an
accuracy of 99.83\%, leaving only 24 misclassified sentences: sixteen from
group A, three from group D, and five from group E. We included most of these instances in a manual annotation experiment providing twelve instances to several native speakers of these languages and asking them to assign the correct language or language variety of each text. 

In our experimental setting, we made sure that the annotators were not exclusively speakers of one of the languages or varieties of the group. However, a perfect balance in participation between languages was very difficult to be obtained. Within a group of two (or three) languages or varieties, native speaker's perception of whether a given text belongs to his own language or not may vary according to the person's own language. We discuss this issue later in this section taking group D (Brazilian and European Portuguese) as an example. 

As to the participants, for group A we asked six annotators (one Bosnian, two Croatians, and three Serbians); for group D we had ten annotators participating (eight Brazilians and two Portuguese); and finally for Group E, seventeen annotators participated (fifteen Argentinians and two Spaniards). In Table \ref{tab:manual} we report the percentage of times that instances 1 to 12 from each group were correctly annotated by the native speakers along with the maximum, minimum, and mean performance of the annotators.

\begin{table}[!ht]
\centering
\scalebox{1.0}{
\begin{tabular}{lccc}
\hline
\bf Group & \bf A & \bf D & \bf E \\
\hline
Classes & 3 & 2 & 2 \\
Annotators	& 6 & 10 & 17 \\
\hline
$\%$ Correct Inst. ID 1 & 16.66 & 30.00 & 35.30 \\
$\%$ Correct Inst. ID 2 & 0 & 80.00 & 29.41 \\
$\%$ Correct Inst. ID 3 & 16.66 & 80.00 & 58.82 \\
$\%$ Correct Inst. ID 4 & 0 & 60.00 & 47.05 \\
$\%$ Correct Inst. ID 5 & 33.33 & 50.00 & 35.29 \\
$\%$ Correct Inst. ID 6 & 0 & 30.00 & 76.47 \\
$\%$ Correct Inst. ID 7 & 0 & 70.00 & 76.47 \\
$\%$ Correct Inst. ID 8 & 66.66 & 80.00 & 11.76 \\
$\%$ Correct Inst. ID 9 & 16.66 & 90.00 & 35.29 \\
$\%$ Correct Inst. ID 10 & 16.66 & 60.00 & 100 \\
$\%$ Correct Inst. ID 11 & 33.33 & 80.00 & 82.35 \\
$\%$ Correct Inst. ID 12 & 0 & 100 & 70.59 \\
\hline
Best Annotator Accuracy & 25.00 & 91.66 & 83.33 \\
Mean Annotator Accuracy & 16.66 & 67.50 & 54.90 \\
Worst Annotator Accuracy & 8.33 & 25.00 & 16.66 \\
\hline
\end{tabular}
}
\caption{Manual Evaluation Results for Groups A, D and E.}
\label{tab:manual}
\end{table}

Group A (Bosnian, Croatian and Serbian) was the most difficult for the annotators. The language of five out of twelve instances was not correctly assigned by any of the six annotators. This is firstly explained because group A is the only group containing three languages. However, we also noted that classifiers showed very high degree of confusion when discriminating between Bosnian and Croatian texts. All instances from this group misclassified by the oracle, and therefore included in this experiment, were either Bosnian or Croatian. The shared task organizers and our paper assume that all gold labels are correct. It might be the case, however, that a few labels in group A contain errors. In case there are incorrect reference labels, these errors would surely negatively impact both classifiers' and humans' performance. This is a possibility that we cannot confirm nor disregard. For pragmatic reasons we rely on the sources that were compiled for the DSLCC.\footnote{As mentioned in Section 2.1. incorrect tags in the gold data were present in the American and British English dataset of the DSLCC v 1.0 (see \newcite{zampieri2014vardial} for a discussion).}

Albeit still challenging, the task proved to be more feasible for group E than for group A. The average performance of the annotators was slightly above the 50\% baseline. One of the seventeen annotators was able to correctly assign the language of the texts ten out of twelve times, achieving 83.33\% accuracy. For all the three groups studied, the only two cases in which all annotators correctly assigned the language of an instance were group group D ID 12 and group E ID 10. To exemplify, the latter is the following Peninsular Spanish text:

\enumsentence{Entonces lo entiendo todo. La prensa no tiene razón. No estamos en guerra, ni falta que hace. Esta especie de bronca continua es una ilusión, una imagen grotesca que proyectan los medios, pero no es real. Por el amor de Dios, que no os pase como a mí, que meriendo paellas de orfidales todos los días. En ocasiones veo Truebas, sí, pero mis amigos me ponen en mi sitio. No nos dejemos engañar.}

The best results were obtained by the annotators of group D (Brazilian and European Portuguese). The average performance of the annotators was 17.50 percentage points above the baseline. This corroborates the findings of \newcite{zampieriandgebre12} who showed that due to differences in spelling and lexical variation, Brazilian and Portuguese texts can be discriminated automatically with almost perfect performance (researchers report 99.8\% accuracy). 

One interesting finding of this experiment is that the competence of identifying whether a given text comes from Brazilian or European Portuguese does not seem to be the same for speakers of these two varieties. Performance highly depends on the lexical variation included. To exemplify, next we present the text included as group D ID 3:

\enumsentence{O médio Bruno Neves tem 25 anos e jogou a última época no Grêmio, do Brasil, tendo também alinhado já no Fluminense e no Cruzeiro.}

The eight Brazilians were unanimous in assigning this sentence as European Portuguese, whereas the two European Portuguese speakers assigned them as Brazilian Portuguese.

This is a particularly interesting example because it talks about three Brazilian football clubs: {\em Grêmio, Fluminense}, and {\em Cruzeiro}. Thematically, this sentence is much more likely to be published in a Brazilian newspaper than in a Portuguese one, and due to the influence of these named entities the sentence was misclassified. However, it features terms that are exclusively used in European Portuguese  such as {\em médio} (BR: {\em meia} or {\em meio-campista}, EN: {\em midfielder}) and {\em época} (BR: {\em temporada}, EN: {\em season}).\footnote{See \newcite{soares10} for a study on lexical variation involving Brazilian and European Portuguese football terms.} For Brazilians, these words were probably an indication that the text was not written in Brazilian Portuguese. On the other hand, European Portuguese speakers who were not aware that these two words are not used in Brazil, probably were influenced by the club names to assume that the sentence is from Brazil.

To sum up, manual annotation for this task is by no means trivial which also explains the difficulty that algorithms have in discriminating between similar languages. We confirm that named entities play an important role in this task and that they can influence not only the performance of algorithms but also the performance of human annotators. Finally, a general tendency we observed is that it is easier to identify an instance that is not from the speaker's own language than the opposite. Our results indicate that humans are better in telling what is not a text written in their own language or variety than telling what it is. We would like to investigate this phenomenon in the future using more annotated data.



\section{Conclusion and Future Work}

This paper presents a comprehensive evaluation of state-of-the-art language identification systems trained to recognized similar languages and language varieties using the results of the first two DSL shared tasks. We evaluate the progress made from one edition of the shared task to the next. Using plurality voting and oracle, we estimate an upper bound on the achievable performance, and identify some particularly challenging sentences. We show learning curves that help us identify how the task is learned and which groups of languages may need more attention. 

Finally, we propose an experiment with native speakers of the three most challenging language groups, group A (Bosnian, Croatian, and Serbian), group D (Brazilian and European Portuguese), and group E (Argentinian and Peninsular Spanish). Our results suggest that humans also find it difficult  discriminating between similar languages and language varieties. In future work we would like to investigate human performance in this task focusing on two aspects: 1) how native speakers identify language variation; 2) which words, expressions or syntactic structures are the most discriminating features of a given language or variety according to the speakers of that language. Both of these aspects will provide us new insights into language variation that can be used for linguistic analysis as well as to improve computational methods to discriminate between similar languages.

\section*{Acknowledgements}

We would like to thank the human annotators who participated in the manual evaluation experiment. We also thank the DSL shared task organizers for providing the data we used in this paper.



\section*{Bibliographical References}
\label{main:ref}

\bibliographystyle{lrec2016}
\bibliography{bibliography}

\begin{thebibliography}{}

\bibitem[\protect\citename{\'{A}cs \bgroup et al.\egroup }2015]{acs2015dsl}
\'{A}cs, J., Grad-Gyenge, L., and de~Rezende~Oliveira, T. B.~R.
\newblock (2015).
\newblock A two-level classifier for discriminating similar languages.
\newblock In {\em Proceedings of the LT4VarDial Workshop}.

\bibitem[\protect\citename{Baldwin and Lui}2010]{baldwin10}
Baldwin, T. and Lui, M.
\newblock (2010).
\newblock Multilingual language identification: {ALTW} 2010 shared task data.
\newblock In {\em Proceedings of Australasian Language Technology Workshop}.

\bibitem[\protect\citename{Bobicev}2015]{bobicev2015dsl}
Bobicev, V.
\newblock (2015).
\newblock Discriminating between similar languages using ppm.
\newblock In {\em Proceedings of the LT4VarDial Workshop}.

\bibitem[\protect\citename{Brown}2013]{brown13}
Brown, R.
\newblock (2013).
\newblock Selecting and weighting n-grams to identify 1100 languages.
\newblock In {\em Proceedings of TSD}.

\bibitem[\protect\citename{Brown}2014]{brown14}
Brown, R.~D.
\newblock (2014).
\newblock Non-linear mapping for improved identification of 1300+ languages.
\newblock In {\em Proceedings of EMNLP}.

\bibitem[\protect\citename{Ciobanu and Dinu}2016]{ciobanu2016}
Ciobanu, A.~M. and Dinu, L.~P.
\newblock (2016).
\newblock A computational perspective on {R}omanian dialects.
\newblock In {\em Proceedings of LREC}.

\bibitem[\protect\citename{Dunning}1994]{dunning94}
Dunning, T.
\newblock (1994).
\newblock Statistical identification of language.
\newblock Technical report, Computing Research Lab - New Mexico State
  University.

\bibitem[\protect\citename{Elfardy and Diab}2014]{elfardy13}
Elfardy, H. and Diab, M.~T.
\newblock (2014).
\newblock Sentence level dialect identification in {A}rabic.
\newblock In {\em Proceedings of ACL}.

\bibitem[\protect\citename{Fabra-Boluda \bgroup et al.\egroup
  }2015]{boluda2015dsl}
Fabra-Boluda, R., Rangel, F., and Rosso, P.
\newblock (2015).
\newblock {NLEL} {UPV} autoritas participation at {D}iscrimination between
  {S}imilar {L}anguages {(DSL)} 2015 shared task.
\newblock In {\em Proceedings of the LT4VarDial Workshop}.

\bibitem[\protect\citename{Franco-Salvador \bgroup et al.\egroup
  }2015]{salvador2015dsl}
Franco-Salvador, M., Rosso, P., and Rangel, F.
\newblock (2015).
\newblock Distributed representations of words and documents for discriminating
  similar languages.
\newblock In {\em Proceedings of the LT4VarDial Workshop}.

\bibitem[\protect\citename{Gebre \bgroup et al.\egroup }2013]{gebreetal2013}
Gebre, B.~G., Zampieri, M., Wittenburg, P., and Heskens, T.
\newblock (2013).
\newblock Improving native language identification with tf-idf weighting.
\newblock In {\em Proceedings of the 8th BEA workshop}.

\bibitem[\protect\citename{Goutte and L\'{e}ger}2015]{goutte2015dsl}
Goutte, C. and L\'{e}ger, S.
\newblock (2015).
\newblock Experiments in discriminating similar languages.
\newblock In {\em Proceedings of the LT4VarDial Workshop}.

\bibitem[\protect\citename{Goutte \bgroup et al.\egroup }2014]{goutteetal2014}
Goutte, C., L\'{e}ger, S., and Carpuat, M.
\newblock (2014).
\newblock The {NRC} system for discriminating similar languages.
\newblock In {\em Proceedings of the VarDial Workshop}.

\bibitem[\protect\citename{Grefenstette}1995]{grefenstette95}
Grefenstette, G.
\newblock (1995).
\newblock Comparing two language identification schemes.
\newblock In {\em Proceedings of JADT 1995, 3rd International Conference on
  Statistical Analysis of Textual Data}, Rome.

\bibitem[\protect\citename{Huang and Lee}2008]{huang08}
Huang, C. and Lee, L.
\newblock (2008).
\newblock Contrastive approach towards text source classification based on
  top-bag-of-word similarity.
\newblock In {\em Proceedings of PACLIC}.

\bibitem[\protect\citename{Jauhiainen \bgroup et al.\egroup
  }2015]{jauhiainen2015dsl}
Jauhiainen, T., Jauhiainen, H., and Lind\'{e}n, K.
\newblock (2015).
\newblock Discriminating similar languages with token-based backoff.
\newblock In {\em Proceedings of the LT4VarDial Workshop}.

\bibitem[\protect\citename{King \bgroup et al.\egroup }2014]{kingetal2014}
King, B., Radev, D., and Abney, S.
\newblock (2014).
\newblock Experiments in sentence language identification with groups of
  similar languages.
\newblock In {\em Proceedings of the VarDial Workshop}.

\bibitem[\protect\citename{Kuncheva \bgroup et al.\egroup
  }2001]{kuncheva2001decision}
Kuncheva, L.~I., Bezdek, J.~C., and Duin, R.~P.
\newblock (2001).
\newblock Decision templates for multiple classifier fusion: an experimental
  comparison.
\newblock {\em Pattern Recognition}, 34(2):299--314.

\bibitem[\protect\citename{Ljube{\v{s}}i{\'c} and
  Kranj{\v{c}}i{\'c}}2015]{ljubesic15}
Ljube{\v{s}}i{\'c}, N. and Kranj{\v{c}}i{\'c}, D.
\newblock (2015).
\newblock Discriminating between closely related languages on twitter.
\newblock {\em Informatica}, 39(1).

\bibitem[\protect\citename{Ljube{\v s}i{\'c} \bgroup et al.\egroup
  }2007]{ljubesic07}
Ljube{\v s}i{\'c}, N., Mikelic, N., and Boras, D.
\newblock (2007).
\newblock Language identification: How to distinguish similar languages?
\newblock In {\em Proceedings of the 29th International Conference on
  Information Technology Interfaces}.

\bibitem[\protect\citename{Lui and Cook}2013]{lui13}
Lui, M. and Cook, P.
\newblock (2013).
\newblock Classifying {E}nglish documents by national dialect.
\newblock In {\em Proceedings of Australasian Language Technology Workshop}.

\bibitem[\protect\citename{Lui \bgroup et al.\egroup }2014a]{lui2014automatic}
Lui, M., Lau, J.~H., and Baldwin, T.
\newblock (2014a).
\newblock Automatic detection and language identification of multilingual
  documents.
\newblock {\em Transactions of the Association for Computational Linguistics},
  2:27--40.

\bibitem[\protect\citename{Lui \bgroup et al.\egroup }2014b]{luietal2014}
Lui, M., Letcher, N., Adams, O., Duong, L., Cook, P., and Baldwin, T.
\newblock (2014b).
\newblock Exploring methods and resources for discriminating similar languages.
\newblock In {\em Proceedings of VarDial}.

\bibitem[\protect\citename{Maier and G{\'o}mez-Rodr{\i}guez}2014]{maier2014}
Maier, W. and G{\'o}mez-Rodr{\i}guez, C.
\newblock (2014).
\newblock Language variety identification in {S}panish tweets.
\newblock In {\em Proceedings of the LT4CloseLang Workshop}.

\bibitem[\protect\citename{Malmasi and Dras}2014]{malmasi-lth}
Malmasi, S. and Dras, M.
\newblock (2014).
\newblock {Language Transfer Hypotheses with Linear SVM Weights}.
\newblock In {\em {Proceedings of EMNLP}}.

\bibitem[\protect\citename{Malmasi and Dras}2015a]{malmasi-dras2015}
Malmasi, S. and Dras, M.
\newblock (2015a).
\newblock {Automatic Language Identification for Persian and Dari texts}.
\newblock In {\em Proceedings of PACLING 2015}, pages 59--64.

\bibitem[\protect\citename{Malmasi and Dras}2015b]{malmasi2015dsl}
Malmasi, S. and Dras, M.
\newblock (2015b).
\newblock Language identification using classifier ensembles.
\newblock In {\em Proceedings of the LT4VarDial Workshop}.

\bibitem[\protect\citename{Malmasi and Dras}2015c]{malmasi-mnli}
Malmasi, S. and Dras, M.
\newblock (2015c).
\newblock {Multilingual Native Language Identification}.
\newblock In {\em Natural Language Engineering}.

\bibitem[\protect\citename{Malmasi \bgroup et al.\egroup
  }2015a]{malmasietal2015}
Malmasi, S., Refaee, E., and Dras, M.
\newblock (2015a).
\newblock {Arabic Dialect Identification using a Parallel Multidialectal
  Corpus}.
\newblock In {\em Proceedings of PACLING 2015}, pages 209--217, Bali,
  Indonesia, May.

\bibitem[\protect\citename{Malmasi \bgroup et al.\egroup
  }2015b]{malmasi2015oracle}
Malmasi, S., Tetreault, J., and Dras, M.
\newblock (2015b).
\newblock Oracle and human baselines for native language identification.
\newblock In {\em Proceedings of the BEA workshop}.

\bibitem[\protect\citename{Manning \bgroup et al.\egroup
  }2008]{manning:2008:eval}
Manning, C.~D., Raghavan, P., and Sch{\"u}tze, H.
\newblock (2008).
\newblock Evaluation in information retrieval.
\newblock In {\em Introduction to Information Retrieval}, pages 151--175.
  Cambridge university press Cambridge.

\bibitem[\protect\citename{Polikar}2006]{polikar2006ensemble}
Polikar, R.
\newblock (2006).
\newblock Ensemble based systems in decision making.
\newblock {\em Circuits and systems magazine, IEEE}, 6(3):21--45.

\bibitem[\protect\citename{Porta and Sancho}2014]{portaandsancho2014}
Porta, J. and Sancho, J.-L.
\newblock (2014).
\newblock Using maximum entropy models to discriminate between similar
  languages and varieties.
\newblock In {\em Proceedings of the 1st Workshop on Applying NLP Tools to
  Similar Languages, Varieties and Dialects (VarDial)}, Dublin, Ireland.

\bibitem[\protect\citename{Purver}2014]{purver2014}
Purver, M.
\newblock (2014).
\newblock A simple baseline for discriminating similar language.
\newblock In {\em Proceedings of the 1st Workshop on Applying NLP Tools to
  Similar Languages, Varieties and Dialects (VarDial)}, Dublin, Ireland.

\bibitem[\protect\citename{{Ranaivo-Malan\c{c}on}}2006]{ranaivo06}
{Ranaivo-Malan\c{c}on}, B.
\newblock (2006).
\newblock Automatic identification of close languages - case study: Malay and
  {I}ndonesian.
\newblock {\em ECTI Transactions on Computer and Information Technology},
  2:126--134.

\bibitem[\protect\citename{Sim{\~o}es \bgroup et al.\egroup }2014]{simoes2014}
Sim{\~o}es, A., Almeida, J.~J., and Byers, S.~D.
\newblock (2014).
\newblock Language identification: a neural network approach.
\newblock {\em Proceedings of Slate}.

\bibitem[\protect\citename{{Soares da Silva}}2010]{soares10}
{Soares da Silva}, A.
\newblock (2010).
\newblock Measuring and parameterizing lexical convergence and divergence
  between {E}uropean and {B}razilian {P}ortuguese: endo/exogeneousness and
  foreign and normative influence.
\newblock {\em Advances in Cognitive Sociolinguistics}.

\bibitem[\protect\citename{Solorio \bgroup et al.\egroup }2014]{solorio14}
Solorio, T., Blair, E., Maharjan, S., Bethard, S., Diab, M., Ghoneim, M.,
  Hawwari, A., AlGhamdi, F., Hirschberg, J., Chang, A., and Fung, P.
\newblock (2014).
\newblock Overview for the first shared task on language identification in
  code-switched data.
\newblock In {\em Proceedings of the First Workshop on Computational Approaches
  to Code Switching}.

\bibitem[\protect\citename{Tan \bgroup et al.\egroup }2014]{tan14}
Tan, L., Zampieri, M., Ljube{\v s}i{\'c}, N., and Tiedemann, J.
\newblock (2014).
\newblock Merging comparable data sources for the discrimination of similar
  languages: The {DSL} corpus collection.
\newblock In {\em Proceedings of The BUCC Workshop}.

\bibitem[\protect\citename{Tiedemann and Ljube{\v s}i{\'c}}2012]{tiedemann12}
Tiedemann, J. and Ljube{\v s}i{\'c}, N.
\newblock (2012).
\newblock Efficient discrimination between closely related languages.
\newblock In {\em Proceedings of COLING}.

\bibitem[\protect\citename{Tillmann \bgroup et al.\egroup
  }2014]{tillmann-mansour-alonaizan:2014:VarDial}
Tillmann, C., Mansour, S., and Al-Onaizan, Y.
\newblock (2014).
\newblock Improved sentence-level {A}rabic dialect classification.
\newblock In {\em Proceedings of the VarDial Workshop}, pages 110--119, Dublin,
  Ireland, August.

\bibitem[\protect\citename{Zaidan and Callison-Burch}2014]{zaidan14}
Zaidan, O.~F. and Callison-Burch, C.
\newblock (2014).
\newblock Arabic dialect identification.
\newblock {\em Computational Linguistics}.

\bibitem[\protect\citename{Zampieri and Gebre}2012]{zampieriandgebre12}
Zampieri, M. and Gebre, B.~G.
\newblock (2012).
\newblock Automatic identification of language varieties: The case of
  {P}ortuguese.
\newblock In {\em Proceedings of KONVENS}.

\bibitem[\protect\citename{Zampieri \bgroup et al.\egroup
  }2013]{zampierietal13}
Zampieri, M., Gebre, B.~G., and Diwersy, S.
\newblock (2013).
\newblock N-gram language models and {POS} distribution for the identification
  of {S}panish varieties.
\newblock In {\em Proceedings of TALN}.

\bibitem[\protect\citename{Zampieri \bgroup et al.\egroup
  }2014]{zampieri2014vardial}
Zampieri, M., Tan, L., Ljube\v{s}i\'{c}, N., and Tiedemann, J.
\newblock (2014).
\newblock A report on the {DSL} shared task 2014.
\newblock In {\em Proceedings of the VarDial Workshop}.

\bibitem[\protect\citename{Zampieri \bgroup et al.\egroup
  }2015a]{zampieri2015dsl}
Zampieri, M., Gebre, B.~G., Costa, H., and {van Genabith}, J.
\newblock (2015a).
\newblock Comparing approaches to the identification of similar languages.
\newblock In {\em Proceedings of the LT4VarDial Workshop}.

\bibitem[\protect\citename{Zampieri \bgroup et al.\egroup
  }2015b]{zampieri15dsl}
Zampieri, M., Tan, L., Ljube\v{s}i\'{c}, N., Tiedemann, J., and Nakov, P.
\newblock (2015b).
\newblock Overview of the {DSL} shared task 2015.
\newblock In {\em Proceedings of LT4VarDial}.

\bibitem[\protect\citename{Zubiaga \bgroup et al.\egroup }2014]{zubiaga14}
Zubiaga, A., San~Vicente, I., Gamallo, P., Pichel, J.~R., Alegria, I.,
  Aranberri, N., Ezeiza, A., and Fresno, V.
\newblock (2014).
\newblock Overview of {TweetLID}: {Tweet} language identification at {SEPLN
  2014}.
\newblock In {\em Proceedings of SEPLN}.

\bibitem[\protect\citename{Zubiaga \bgroup et al.\egroup
  }2015]{zubiaga2015tweetlid}
Zubiaga, A., San~Vicente, I., Gamallo, P., Pichel, J.~R., Alegria, I.,
  Aranberri, N., Ezeiza, A., and Fresno, V.
\newblock (2015).
\newblock Tweetlid: a benchmark for tweet language identification.
\newblock {\em Language Resources and Evaluation}, pages 1--38.

\end{thebibliography}

\end{document}